\documentclass[letter, 10pt, conference]{ieeeconf}      

\usepackage{graphicx,balance}
\usepackage{epstopdf}

\usepackage[sort,compress]{cite}

\usepackage{amsmath,amssymb,amsthm}
\usepackage{algorithm,algpseudocode}

\usepackage{color}

\usepackage{subfigure}

\usepackage{tikz}
\usetikzlibrary{shapes,arrows,automata,calc,trees,positioning,fit,shapes,calc,patterns,3d}
\usepackage{pgfplots}
\usepgfplotslibrary{fillbetween,groupplots}

\usetikzlibrary{external}
\tikzsetexternalprefix{externalize/}
\tikzset{external/optimize=false}

\IEEEoverridecommandlockouts
\theoremstyle{plain}

\theoremstyle{definition}

\theoremstyle{remark}

\DeclareMathOperator*{\argmin}{argmin}
\DeclareMathOperator*{\argmax}{argmax}

\author{Miguel Calvo-Fullana and Jonathan P.\ How
\thanks{Work was supported in part by Lockheed Martin Corporation and by ONR under BRC award N000141712072. Authors at Massachusetts Institute of Technology, Cambridge, MA, USA. e-mail: \{cfullana,jhow\}@mit.edu.
}
}

\title{\LARGE \bf
Mission-Aware Value of Information Censoring for Distributed Filtering
}

\begin{document}

\maketitle

\begin{abstract}
In this paper, we study the problem of distributed estimation with an emphasis on communication-efficiency. The proposed algorithm is based on a windowed maximum a posteriori (MAP) estimation problem, wherein each agent in the network locally computes a Kalman-like filter estimate that approximates the centralized MAP solution. Information sharing among agents is restricted to their neighbors only, with guarantees on overall estimate consistency provided via logarithmic opinion pooling. The problem is efficiently distributed using the alternating direction method of multipliers (ADMM), whose overall communication usage is further reduced by a value of information (VoI) censoring mechanism, wherein agents only transmit their primal-dual iterates when deemed valuable to do so. The proposed censoring mechanism is mission-aware, enabling a globally efficient use of communication resources while guaranteeing possibly different local estimation requirements. To illustrate the validity of the approach we perform simulations in a target tracking scenario. 
\end{abstract}

\section{Introduction}

Distributed estimation plays an important role in many systems and domains. The desire for distributed operation stems from many different sources. In some cases, centralization is simply not possible, and in other cases, distributed operation is strongly preferred, whether it be due to issues related to scalability, robustness to single point of failure, or any other reasons. Importantly, decentralized operation introduces a new set of concerns into the estimation problem, as issues such as distributed computation or communication efficiency take a prominent role in algorithm design.

Under the traditional assumption of linear-Gaussian dynamics, the Kalman filter \cite{kalman1960new} provides an optimal centralized solution to the sequential estimation problem. Due to this, distributed variants of the Kalman filter have received considerable attention in the literature. Chief among them is the consensus Kalman filter \cite{olfati2009kalman} and a large body of related variants \cite{meng2014optimality,liu2015event,ouimet2018cooperative}. A key challenge among these methods is the fact that consensus is only achieved to the mean of the estimate, with covariances possibly diverging among different nodes in the network \cite{kamgarpour2008convergence}. Issues that are usually alleviated by combining these methods with covariance intersection rules \cite{julier1997non,chen2002estimation}. Furthermore, these concerns also exist and have been studied for more general non-linear and non-Gaussian systems. In the general case, the optimal filtering process is provided by the Bayes' filter \cite{chen2003bayesian}, of which distributed versions have been obtained via logarithmic opinion pooling \cite{bandyopadhyay2014distributed,bandyopadhyay2018distributed}.

Another major concern of distributed estimation methods is communication efficiency. In general, approximating the centralized estimate requires flooding information across the network \cite{olfati2007distributed}, which can be prohibitively expensive in communication resources. Attempts to mitigate this problem have led to algorithms based on a single round of communication among neighbors \cite{kamal2013information,das2016consensus} and further reductions have been achieved by the use of censoring and communication-triggered estimation mechanisms \cite{nowzari2019event,ouimet2018cooperative,liu2015event,meng2014optimality}.

This work presents a communication-efficient distributed filtering algorithm. We formulate a maximum a posteriori (MAP) estimation problem, which, to solve it in a sequential manner while staying close to the batch estimate, we reformulate as a rolling window tracking problem. Motivated by \cite{shorinwa2020distributed}, we distribute the problem using the alternating direction method of multipliers (ADMM) \cite{boyd2011distributed}. Under linear-Gaussian assumptions, the iterates of the algorithm can be computed in closed form, resulting in a variant of the Kalman filter. The proposed distributed algorithm approximates the centralized solution, addressing the issue of covariance discrepancy among network nodes via the use of logarithmic opinion pooling. This method aggregates information with neighborhood-only communication, providing guarantees in estimate consistency. 

In order to further reduce the communication overhead, we introduce a censoring mechanism based on Value of Information (VoI) \cite{mu2014efficient}, which we have previously applied with success to distributed filtering \cite{calvo2022distributed}. At a given time instance, this censoring method compares by way of the Kullback-Leibler divergence, a node's local information and the incoming information from its measurements and neighbors, transmitting when information is deemed sufficiently valuable. In contrast to our previous work \cite{calvo2022distributed}, in which the state of the target determines the VoI, in this work the role of information pertains to the primal-dual variables. This enables the VoI censoring mechanism to be further enhanced by the introduction of mission-awareness, accounting for individual requirements that must be satisfied by agents to operate successfully. These local mission requirements are introduced in the form of estimation constraints distributed by the ADMM mechanism. This allows for different agents in the network to guarantee different estimation requirements, while making an efficient use of communication resources across the network. Finally, we test the proposed algorithm in a simulated target tracking scenario, where a set of heterogeneous nodes with different sensing capabilities attempt to jointly estimate the trajectory of a target.  Compared to our previous work \cite{calvo2022distributed}, these results show that the proposed method is capable of achieving lower estimation error while being communication-efficient and guaranteeing node-specific mission requirements.

\section{Problem Formulation}

We begin by considering a discrete-time dynamic process with state $\mathbf{x}_k \in \mathbb{R}^n$ at a given $k$-th time instant. This process evolves following linear-Gaussian dynamics given by
\begin{align}
\label{eq:dynamics}
\mathbf{x}_{k+1}=\mathbf{A}_k \mathbf{x}_k + \mathbf{w}_k,
\end{align}
with dynamics matrix $\mathbf{A}_k \in \mathbb{R}^{n \times n}$ and	independent and identically distributed (i.i.d.) additive Gaussian noise $\mathbf{w}_k \sim \mathcal{N}(\mathbf{0},\mathbf{Q}_k)$. We consider a set of $N$ sensors performing an estimation task. The  $i$-th sensor at time $k$ obtains an observation vector $\mathbf{z}^i_{k} \in \mathbb{R}^{m_i}$ given by
\begin{align}
\label{eq:measurements}
\mathbf{z}^i_{k}=\mathbf{H}^i_{k} \mathbf{x}_k + \mathbf{v}^i_{k},
\end{align}
with measurement matrix $\mathbf{H}^i_{k}  \in \mathbb{R}^{m_i \times n	}$ and measurement noise $\mathbf{v}^i_{k} \in \mathbb{R}^{m_i}$ distributed according to $\mathbf{v}^i_{k} \sim \mathcal{N}(0,\mathbf{R}_{k}^{i})$. First, for simplicity, we will formulate the centralized estimation problem. To this end, we can concatenate all the measurements to obtain the joint centralized observation
\begin{align}
\label{eq:measurements_centralized}
\mathbf{z}_{k}=\mathbf{H}_{k} \mathbf{x}_k + \mathbf{v}_{k},
\end{align}
where $\mathbf{z}_{k}$, $\mathbf{H}_{k}$, and $\mathbf{v}_{k}$, correspond to the column-wise concatenations of the specific $i$-th agent counterparts.

\subsection{Formulating the Centralized MAP Estimation Problem}

In order to obtain a centralized estimate, we infer the state based on the joint centralized observations, and a prior $\bar{\mathbf{x}}_0$ with covariance $\bar{\mathbf{P}}_0$. We can then formulate the maximum a posteriori (MAP) batch estimate as follows
\begin{align}
\hat{\mathbf{x}}^{\text{MAP}}_{0:k}=\argmax_{\mathbf{x}_{0:k}} p(\mathbf{x}_k ~| \bar{\mathbf{x}}_0, \mathbf{z}_{1:k})
\end{align}
where we have introduced the notation $\mathbf{x}_{0:k}$ to denote the state trajectory over the time horizon $[0,\ldots,k]$ and equivalently with $\mathbf{z}_{1:k}$ for the set of observations $\{\mathbf{z}_1,\ldots,\mathbf{z}_k\}$. Now, using Bayes' rule we can rewrite this problem as
\begin{align}
\label{eq:max_MAP_problem}
\hat{\mathbf{x}}^{\text{MAP}}_{0:k}\! = \!\argmax_{\mathbf{x}_{0:k}} 
p(\mathbf{x}_0~| \bar{\mathbf{x}}_0) 
\prod_{l=1}^{k}
p(\mathbf{x}_{l} | \mathbf{x}_{l-1})
\prod_{l=1}^{k}
p(\mathbf{z}_l~| \mathbf{x}_l).
\end{align}
Under the linear-Gaussian model \eqref{eq:dynamics}--\eqref{eq:measurements_centralized}, the previous maximization can be solved by forming a system of linear equations. In practice, we might however opt to solve \eqref{eq:max_MAP_problem} in a sequential manner. The Kalman filter does so, obtaining the estimate $\hat{\mathbf{x}}_{k}$ by conditioning on the prior estimate $\bar{\mathbf{x}}_{k-1}$. In any case, the Kalman filter only replicates the solution of the batch estimate \eqref{eq:max_MAP_problem} at the final time instance $k$. To ameliorate this shortcoming, we include a rolling window into the estimation problem, which has shown success in practice \cite{shorinwa2020distributed}. We can then write the rolling window tracking (RWT) estimate over time horizon $H$ as 
\begin{align}
\label{eq:max_RWT_problem}
\hat{\mathbf{x}}^{\text{RWT}}_{k-H:k} \!\!= \!\! \argmax_{\mathbf{x}_{k-H:k}} 
p(\mathbf{x}_{k-H:k-1}| \bar{\mathbf{x}}_{k-H:k-1}) 
p(\mathbf{x}_{k}| \mathbf{x}_{k-1})
p(\mathbf{z}_k | \mathbf{x}_k) 
\end{align}
Then, under the linear-Gaussian model \eqref{eq:dynamics}--\eqref{eq:measurements_centralized}, we can define the cost
\begin{align}
J(\mathbf{x}_{k-H:k}) & =
\|\mathbf{x}_k - \mathbf{A}_{k-1} \mathbf{x}_{k-1} \|^2_{\mathbf{Q}_{k-1}^{-1}}
+ \| \mathbf{z}_k - \mathbf{H}_k \mathbf{x}_k \|^2_{\mathbf{R}_k^{-1}}\nonumber \\
& + \| \mathbf{x}_{k-H:k-1} - \bar{\mathbf{x}}_{k-H:k-1} \|^2_{\bar{\mathbf{P}}_{k-H:k-1}^{-1}} 
\end{align}
and write the estimation task in the optimization problem form as simply the minimization of this cost, \cite{chen2003bayesian}
\begin{align}
\label{eq:opt_centralized}
   \hat{\mathbf{x}}^{\text{RWT}}_{k-H:k}=\argmin_{\mathbf{x}_{k-H:k}}  J(\mathbf{x}_{k-H:k}).
\end{align}

\section{Distributed Estimation Problem}
\label{sec:distributed}

The problem \eqref{eq:opt_centralized} introduced in the previous section is a centralized problem, which requires centralized observations \eqref{eq:measurements_centralized}. In practice, we have a network of agents, each obtaining different observations \eqref{eq:measurements} of the target. In order to perform the estimation task, the sensors must communicate with each other. They do so over a communication network given by the network graph  $\mathcal{G}=(\mathcal{N},\mathcal{E})$, with $\mathcal{N}$ being the set of $N$ nodes in the network and $\mathcal{E} \subseteq \mathcal{N} \times \mathcal{N}$ the set of communication links, such that if node $i$ is capable of communicating with node $j$, we have $(i,j) \in \mathcal{E}$. Furthermore, we denote the neighborhood of node $i$ by the set $\mathcal{N}_i=\{j | (i,j)\in \mathcal{E}\}$. 

Since communicating all measurements among all nodes in order to solve the centralized problem is prohibitively expensive in communication resources, we resort to a distributed formulation. We obtain a distributed formulation of the problem via the introduction of consensus constraints
\begin{subequations}
\label{eq:opt_distributed}
\begin{align}
   \underset{\mathbf{x}^i_{k-H:k}}{\text{minimize}}  \quad   & \sum_{i=1}^{N} J^{i} (\mathbf{x}^i_{k-H:k})   \\
   \text{subject to}
		\quad & \mathbf{x}^i_{k-H:k} = \mathbf{t}^{ij} \quad \forall j \in \mathcal{N}_i, \forall i \in \mathcal{N} \label{eq:opt_dist_c1}\\
		\quad & \mathbf{x}^j_{k-H:k} = \mathbf{t}^{ij} \quad \forall j \in \mathcal{N}_i, \forall i \in \mathcal{N} \label{eq:opt_dist_c2}
\end{align}
\end{subequations}
where now $\mathbf{x}^i_{k-H:k}$ denotes the rolling window estimate at the $i$-th agent, and the auxiliary variables $\mathbf{t}^{ij}$ have been introduced to enable the distributed operation of the problem. Further, the per-agent cost is defined as
\begin{align}
J^i(\mathbf{x}^{i}_{k-H:k}) & =
\|\mathbf{x}^i_k - \mathbf{A}_{k-1} \mathbf{x}^i_{k-1} \|^2_{\mathbf{Q}_{k-1}^{-1}}
+ \| \mathbf{z}^i_k - \mathbf{H}^i_k \mathbf{x}^i_k \|^2_{\mathbf{R}_k^{i,-1}}\nonumber \\
& + \| \mathbf{x}^i_{k-H:k-1} - \bar{\mathbf{x}}^i_{k-H:k-1} \|^2_{\bar{\mathbf{P}}_{k-H:k-1}^{i,-1}}.
\end{align}
This decomposition matches the centralized one if the prior covariance matrices $\bar{\mathbf{P}}_{k-H:k-1}^i$ across all the network satisfy $\bar{\mathbf{P}}_{k-H:k-1}^{-1}=\sum_{i=i}^{N}\bar{\mathbf{P}}_{k-H:k-1}^{i,-1}$, i.e., they are equivalently decomposed. In practice this would require an asymptotic consensus loop on the covariance matrices, which might pose to be prohibitively expensive, as it requires full communication over the whole network at all instances, defeating some of the purpose of a fully distributed formulation. This problem is at the heart of distributed filtering. When desiring to operate in a neighborhood communication only basis, we resort to a single logarithmic opinion pool (LogOP) consensus step \cite{genest1986combining}. Namely, at the $i$-th node of the network, the prior covariance is updated following
\begin{align}
\bar{\mathbf{P}}_{k-H:k}^{i,-1} = \bar{\mathbf{P}}_{k-H:k-1}^{i,-1} + \sum_{j \in \mathcal{N}_i} \bar{\mathbf{P}}_{k-H:k-1}^{j,-1}.
\end{align}
While this will result in a suboptimal solution (when compared to the centralized one), using this covariance update has several desirables properties. Besides being computable with neighborhood-only communication, it guarantees the consistency of the resulting estimates, preventing the estimate belief across the network does from diverging \cite{bandyopadhyay2018distributed}.

We can now proceed to solve the distributed problem \eqref{eq:opt_distributed}. To do so we resort to the alternating direction method of multipliers (ADMM) \cite{boyd2011distributed}, an efficient method to solve problems of this form. We start by constructing the augmented Lagrangian of the problem. Namely,\footnote{With a slight abuse of notation, for simplicity, we drop the window horizon $[k-H:k]$ from the variables' subscript.}
\begin{align}
\label{eq:Lagrangian}
\mathcal{L}_{\rho}& (\mathbf{x},\boldsymbol{\lambda})=
\sum_{i=1}^{N} J_{i} (\mathbf{x}^i) \\
&+\sum_{i \in \mathcal{N}} \sum_{j \in \mathcal{N}_i} \boldsymbol{\mu}_{ij}^T \bigl(  \mathbf{x}^i - \mathbf{t}^{ij}  \bigr)
+ \sum_{i \in \mathcal{N}} \sum_{j \in \mathcal{N}_i} \boldsymbol{\nu}_{ij}^T \bigl( \mathbf{x}^j - \mathbf{t}^{ij} \bigr) \nonumber\\
&+\frac{\rho}{2} \sum_{i \in \mathcal{N}} \sum_{j \in \mathcal{N}_i}  \bigl\| \mathbf{x}^i - \mathbf{t}^{ij}\bigr\|^2
+\frac{\rho}{2} \sum_{i \in \mathcal{N}} \sum_{j \in \mathcal{N}_i}  \bigl\| \mathbf{x}^j - \mathbf{t}^{ij}\bigr\|^2 \nonumber
\end{align}
where $\rho$ is the scalar penalty parameter of the augmented Lagrangian, and we have introduced the dual variables $\boldsymbol{\mu}_{ij}$ and $\boldsymbol{\nu}_{ij}$, associated with constraints \eqref{eq:opt_dist_c1} and \eqref{eq:opt_dist_c2}, respectively. We can aggregate the per-agent dual variables by defining $\boldsymbol{\lambda}_i ^i\triangleq \sum_{j \in \mathcal{N}_i} ( \boldsymbol{\mu}_{ij} + \boldsymbol{\nu}_{ij} )$ \cite{chang2014multi}. The primal update is given by
\begin{align}
\label{eq:primal_min}
\mathbf{x}^i_{k+1}=\argmin_{\mathbf{x^i}}
\biggl(
&J_{i} (\mathbf{x}^i) + \boldsymbol{\lambda}^{i,T}_{k}\mathbf{x}^i \nonumber\\
&+\rho \sum_{j \in \mathcal{N}_i} \bigl\|\mathbf{x}^i - \frac{\mathbf{x}^i_k+\mathbf{x}^j_k}{2}\bigr\|^2
\biggr),
\end{align}
which by substituting the linear-Gaussian model \eqref{eq:dynamics}--\eqref{eq:measurements_centralized}, we can obtain in closed-form
\begin{align}
\label{eq:primal}
\mathbf{x}^i_{k+1}&=\left( \mathbf{C}_k^{i,T} \mathbf{W}_k^{i,-1} \mathbf{C}_k^{i} + 2 N \rho  \mathbf{I}
\right)^{-1} \nonumber\\
&\times \left(
\mathbf{C}_k^{i,T} \mathbf{W}_k^{i,-1} \mathbf{r}^i_k - \boldsymbol{\lambda}^i_{k}
+ \rho \sum_{j \in \mathcal{N}_i} \bigl(\mathbf{x}^i_{k} + \mathbf{x}^j_{k}\bigr)
\right)
\end{align}
and the covariance update is given by
\begin{align}
\label{eq:cov_update}
\mathbf{P}^i_{k+1}=\bigl( \mathbf{C}_k^{i,T} \mathbf{W}_k^{i,-1} \mathbf{C}_k^{i} \bigr)^{-1},
\end{align}
where we have defined the auxiliary rolling window vector $\mathbf{r}_k^i$, and matrices $\mathbf{W}_k^i$ and $\mathbf{C}_k^i$ as
\begin{align}
\mathbf{r}_k^i=
\begin{bmatrix}
\mathbf{0}\\
\mathbf{z}^i_{t}\\
\bar{\mathbf{x}}^i_{k-1}
\end{bmatrix}
&,\enspace
\mathbf{W}_k^i=
\begin{bmatrix}
\mathbf{Q}_{k-1} & \mathbf{0} & \mathbf{0}\\
\mathbf{0} & \mathbf{R}^i_k & \mathbf{0}\\
\mathbf{0} & \mathbf{0} & \bar{\mathbf{P}}^i_{k-1}
\end{bmatrix},
\end{align}
\begin{align}
\mathbf{C}_k^i&=
\begin{bmatrix}
\mathbf{0} & \cdots &-\mathbf{A}_{k-1} & \mathbf{I}\\
\mathbf{0} & \cdots & \mathbf{0} & \mathbf{H}^i_k\\
\mathbf{I} & \cdots & \mathbf{0} & \mathbf{0} \\
\vdots & \ddots & \vdots & \vdots \\
\mathbf{0} & \cdots & \mathbf{I}  & \mathbf{0} 
\end{bmatrix}.
\end{align}
Finally, the dual update at the $i$-th node is simply obtained by performing gradient ascent on the augmented Lagrangian with respect to the dual variables, resulting in the iterate
\begin{align}
\label{eq:dual}
\boldsymbol{\lambda}^i_{k+1}=\boldsymbol{\lambda}^i_{k}+\rho \sum_{j \in \mathcal{N}_i} \bigl(\mathbf{x}^i_{k} - \mathbf{x}^j_{k}\bigr).
\end{align}
Alternating between the primal minimization \eqref{eq:primal} and dual gradient ascent \eqref{eq:dual}, the agents are guaranteed to converge to the optimal solution of the distributed problem \cite{boyd2011distributed}.

\section{Communication Efficiency}

By solving the distributed problem \eqref{eq:opt_distributed} instead of the centralized problem \eqref{eq:opt_centralized} we have avoided the need of sharing all the information among all the nodes in order to solve the centralized problem. While obtaining the same solution as \eqref{eq:opt_centralized} is still a communication expensive problem to solve, we have further reduced the communication needs by using a LogOP aggregation of prior beliefs, avoiding an inner consensus loop on the covariances. Still, we can obtain further reductions in communication use by censoring information and only transmitting when it is valuable to do so. In order to address this, we introduce the concept of value of information.

\subsection{Value of Information}

Value of Information (VoI) \cite{mu2014efficient} consists on quantifying the information acquired at each node and making transmit decisions depending on it. Formally, we define the value of information as the divergence between local and new information
\begin{align}
	D_{\text{KL}}\bigl(\mathcal{N}(\mathbf{y}^{i}_{k+1}, \mathbf{Y}^{i}_{k+1}) ~\|~ \mathcal{N}(\tilde{\mathbf{y}}^{i}_{k+1}, \tilde{\mathbf{Y}}^{i}_{k+1})\bigr) \geq \gamma
\end{align}
where $D_{\text{KL}}(p \| q)= \int p(x) \log \left(\frac{p(x)}{q(x)}\right) dx$ is the Kullback-Leibler (KL) divergence, and $(\mathbf{y}^{i}_{k+1}, \mathbf{Y}^{i}_{k+1})$ are the aggregated primal-dual variables,
\begin{align}
\mathbf{y}^{i}_{k+1}=
\begin{bmatrix}
\mathbf{x}^{i}_{k+1}\\
\boldsymbol{\lambda}^{i}_{k+1} \\
\end{bmatrix}
,\quad
\mathbf{Y}^{i}_{k+1}=
\begin{bmatrix}
\mathbf{P}_{k+1}^{i} & \mathbf{0}\\
\mathbf{0} & \mathbf{1}\\
\end{bmatrix}
\end{align}
The variables $(\tilde{\mathbf{y}}^{i}_{k+1}, \tilde{\mathbf{Y}}^{i}_{k+1})$, correspond to the baseline local values obtained from the last primal-dual iteration \eqref{eq:primal}--\eqref{eq:dual} being updated without new information, either from local measurements or neighbor transmissions, i.e., $\mathcal{N}_i=\emptyset$. The parameter $\gamma$ corresponds to the censoring level. Only for value on information over this level will the agent transmit its variables $(\mathbf{x}^i_{k+1},\mathbf{P}^i_{k+1})$ to its neighbors.

\subsection{Mission-aware Censoring Mechanism}
\label{sec:mission_aware}

In a communication scenario without censoring, all the required information is transmitted at all times and all agents obtain the best possible estimate. In contrast, this is not necessarily the case when using a censoring mechanism, as a trade-off occurs between communication use and resulting estimation error. Thus, if individual nodes have specific estimation needs, they must be explicitly specific. We denote these as \emph{mission requirements}. Agents have a specific mission (i.e., a task that they need to perform) and their ability to accomplish this task is tied to the estimate itself (e.g., requiring the estimation error to be under a certain threshold). To account for this, we introduce the following mission requirement function
\begin{align}
g^i(\mathbf{x}^i) \leq c_i, \quad i \in \mathcal{N}_m 
\end{align}
where $c_i$ is the level of requirement and $\mathcal{N}_m \subseteq \mathcal{N}$ is the subset of nodes that have a mission specified. We can then introduce this into the distributed estimation problem \eqref{eq:opt_distributed}, formulating the mission-aware estimation problem
\begin{subequations}
\label{eq:opt_distributed_mission}
\begin{align}
   \underset{\mathbf{x}^i_{k-H:k}}{\text{minimize}}  \quad   & \sum_{i=1}^{N} J^{i} (\mathbf{x}^i_{k-H:k})   \\
   \text{subject to}
		\quad & \mathbf{x}^i_{k-H:k} = \mathbf{t}^{ij} \quad \forall j \in \mathcal{N}_i, \forall i \in \mathcal{N} \label{eq:opt_dist_mission_c1}\\
		\quad & \mathbf{x}^j_{k-H:k} = \mathbf{t}^{ij} \quad \forall j \in \mathcal{N}_i, \forall i \in \mathcal{N} \label{eq:opt_dist_mission_c2} \\
		\quad & g^i(\mathbf{x}^i_{k-H:k}) \leq c_i \quad \forall i \in \mathcal{N}_m \label{eq:opt_dist_mission_c3}
\end{align}
\end{subequations}
Similarly to Section \ref{sec:distributed}, we can distribute this problem by constructing the mission-aware augmented Lagrangian
\begin{align}
\mathcal{L}_{\rho}& (\mathbf{x},\boldsymbol{\lambda})=
\sum_{i=1}^{N} J^{i} (\mathbf{x}^i) \\
&+\sum_{i \in \mathcal{N}} \sum_{j \in \mathcal{N}_i} \boldsymbol{\mu}_{ij}^T \bigl(  \mathbf{x}^i - \mathbf{t}^{ij}  \bigr)
+ \sum_{i \in \mathcal{N}} \sum_{j \in \mathcal{N}_i} \boldsymbol{\nu}_{ij}^T \bigl( \mathbf{x}^j - \mathbf{t}^{ij} \bigr) \nonumber\\
&+\frac{\rho}{2} \sum_{i \in \mathcal{N}} \sum_{j \in \mathcal{N}_i}  \bigl\| \mathbf{x}^i - \mathbf{t}^{ij}\bigr\|^2
+\frac{\rho}{2} \sum_{i \in \mathcal{N}} \sum_{j \in \mathcal{N}_i}  \bigl\| \mathbf{x}^j - \mathbf{t}^{ij}\bigr\|^2 \nonumber\\
&+ \sum_{i \in \mathcal{N}_m} \Phi^{i,T} \bigl( g^i(\mathbf{x}^i) - c_i\bigr)
+\frac{\rho}{2} \sum_{i \in \mathcal{N}_m}\|g^i(\mathbf{x}^i) - c_i\|^2, \nonumber
\end{align}
where we have introduced the new Lagrange multipliers $\Phi^i$ associated to the mission constraints \eqref{eq:opt_dist_mission_c3}. The process to obtain the primal update follows the same steps as before, where instead of the primal update \eqref{eq:primal_min} we have now a primal minimization which depends on the mission. Namely,
\begin{align}
\mathbf{x}^i_{k+1}&=\argmin_{\mathbf{x^i}}
\biggl(
J_{i} (\mathbf{x}^i)  +\rho \sum_{j \in \mathcal{N}_i} \bigl\|\mathbf{x}^i - \frac{\mathbf{x}^i_k+\mathbf{x}^j_k}{2}\bigr\|^2  \nonumber\\
&+ \boldsymbol{\lambda}^{i,T}_{k}\mathbf{x}^i  
+ \frac{\rho}{2} \|g^i(\mathbf{x}^i) - c_i + \Phi^i_k\|^2 \mathbb{I}\{i \in \mathcal{N}_m\} 
\biggr)
\end{align}
where $\mathbb{I}\{i \in \mathcal{N}_m\}$ denotes the indicator function such that $\mathbb{I}\{i \in \mathcal{N}_m\}=1$ if $i \in \mathcal{N}_m$. The closed-form solution to this minimization will depend on the structure of the mission function $g^i(\mathbf{x}^i)$. In the case that a specific node has no mission specified, the closed-form solution corresponds to \eqref{eq:primal}. Furthermore, the introduction of the new mission constraints to the optimization problem \eqref{eq:opt_distributed_mission}, results in an additional dual variable that is updated by gradient ascent
\begin{align}
\label{eq:dual_req_update}
\Phi^i_{k+1}=\biggl[\Phi^i_{k} + \rho \bigl( g^i(\mathbf{x}_k^i) - c_i \bigr) \biggr]^+.
\end{align}
Additionally, the aggregated primal-dual variables $(\mathbf{y}^{i}_{k+1}, \mathbf{Y}^{i}_{k+1})$ over which the VoI censoring decision is taken are further augmented with the dual variables associated with the mission requirement. Thus,
\begin{align}
\mathbf{y}^{i}_{k+1}=
\begin{bmatrix}
\mathbf{x}^{i}_{k+1}\\
\boldsymbol{\lambda}^{i}_{k+1} \\
\boldsymbol{\Phi}_{k+1}
\end{bmatrix}
,\quad
\mathbf{Y}^{i}_{k+1}=
\begin{bmatrix}
\mathbf{P}_{k+1}^{i} & \mathbf{0} & \mathbf{0}\\
\mathbf{0} & \mathbf{1} & \mathbf{0}\\
\mathbf{0} & \mathbf{0} & \mathbf{1}
\end{bmatrix}.
\end{align}
Finally, in order to guarantee knowledge of mission requirements across the network, a consensus mechanism is utilized. Nodes keep local copies of the dual variables associated to the mission requirements of other nodes, and update them following an average consensus step, 
\begin{align}
\label{eq:dual_req_cons}
\boldsymbol{\Phi}^{i}_{k+1} = \frac{1}{|\mathcal{N}_i|}\sum_{j \in \mathcal{N}_i} \boldsymbol{\Phi}^{j}_{k+1}.
\end{align}
The agents under the mission update the evolution of the requirement via \eqref{eq:dual_req_update} while the rest of the agents in the network receive this information via network consensus \eqref{eq:dual_req_cons}. This ensures that appropriate knowledge is disseminated across the network, ensuring mission-aware VoI censoring decisions. Overall, the resulting algorithm is summarized in Algorithm \ref{alg:algorithm}.

\begin{algorithm}[t]\caption{Mission-aware VoI-censored distributed filter}
\label{alg:algorithm} 
\begin{algorithmic}[1]
 \renewcommand{\algorithmicrequire}{\textbf{Input:}}
 \renewcommand{\algorithmicensure}{\textbf{Output:}}
 \For {$k=0,1\ldots$}
 \If{mission-aware node ($i \in \mathcal{N}_m$)}
 \State Compute primal update:
\begin{align*}
\mathbf{x}^i_{k+1}=\argmin_{\mathbf{x^i}}&
\biggl(
J_{i} (\mathbf{x}^i)  +\rho \sum_{j \in \mathcal{N}_i} \bigl\|\mathbf{x}^i - \frac{\mathbf{x}^i_k+\mathbf{x}^j_k}{2}\bigr\|^2  \nonumber\\
&+ \boldsymbol{\lambda}^{i,T}_{k}\mathbf{x}^i  
+ \frac{\rho}{2} \|g^i(\mathbf{x}^i) - c_i + \Phi^i_k\|^2
\biggr)
\end{align*}
 \State Compute mission dual update:
\begin{align*}
\Phi^i_{k+1}=\biggl[\Phi^i_{k} + \rho \bigl( g^i(\mathbf{x}_k^i) - c_i \bigr) \biggr]^+
\end{align*}
\ElsIf{not mission-aware node ($i \not\in \mathcal{N}_m$)}
 \State Compute primal update:
 \begin{align*}
\mathbf{x}^i_{k+1}&=\left( \mathbf{C}_k^{i,T} \mathbf{W}_k^{i,-1} \mathbf{C}_k^{i} + 2 N \rho  \mathbf{I}
\right)^{-1} \\
&\times \left(
\mathbf{C}_k^{i,T} \mathbf{W}_k^{i,-1} \mathbf{r}^i_k 
- \boldsymbol{\lambda}^i_{k}
+ \rho \sum_{j \in \mathcal{N}_i} \bigl(\mathbf{x}^i_{k} + \mathbf{x}^j_{k}\bigr)
\right)
\end{align*}
\EndIf
 \State Update local covariance:
\begin{align*}
\mathbf{P}^i_{k+1} &=\bigl( \mathbf{C}_k^{i,T} \mathbf{W}_k^{i,-1} \mathbf{C}_k^{i} \bigr)^{-1}	
\end{align*}
 \State Compute dual update:
\begin{align*}
\boldsymbol{\lambda}^i_{k+1}=\boldsymbol{\lambda}^i_{k}+\rho \sum_{j \in \mathcal{N}_i} \bigl(\mathbf{x}^i_{k} - \mathbf{x}^j_{k}\bigr)
\end{align*}
 \State Transmit $(\mathbf{y}^{i}_{k+1}, \mathbf{Y}^{i}_{k+1})$ if VoI:
\begin{align*}
	D_{\text{KL}}\bigl(\mathcal{N}(\mathbf{y}^{i}_{k+1}, \mathbf{Y}^{i}_{k+1}) ~\|~ \mathcal{N}(\tilde{\mathbf{y}}^{i}_{k+1}, \tilde{\mathbf{Y}}^{i}_{k+1})\bigr) \geq \gamma
\end{align*} 
 \State Merge received neighborhood covariances:
 \begin{align*}
\bar{\mathbf{P}}_{k+1}^{i,-1} = \mathbf{P}_{k+1}^{i,-1} + \sum_{j \in \mathcal{N}_i} \mathbf{P}_{k+1}^{j,-1}
\end{align*}
\State Merge received neighborhood duals:
\begin{align*}
\boldsymbol{\Phi}^{i}_{k+1} = \frac{1}{|\mathcal{N}_i|}\sum_{j \in \mathcal{N}_i} \boldsymbol{\Phi}^{j}_{k+1}
\end{align*}
\EndFor
\end{algorithmic}
\end{algorithm}

\section{Numerical Results}
\label{sec:numericals}

\begin{figure}[t]
	\centering
	\includegraphics[scale=1]{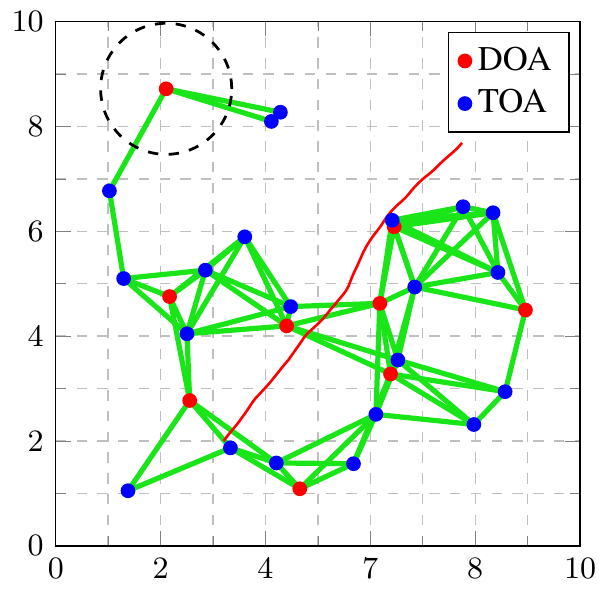} 			
\caption{Map of the evaluation scenario. Axis' distances are shown in Kilometers. The red line corresponds to the target's trajectory. Two types of agents are deployed, equipped with Direction-Of-Arrival (DOA, red) and Time-Of-Arrival (TOA, blue) sensors. All agents have a uniform sensing range of $1.25$km (illustrated by the dashed circle). Underlying network connectivity is shown in green.}
	\label{fig:map}
\end{figure}

We provide now simulations to evaluate the performance of the proposed mission-aware and VoI-censored distributed estimation mechanism. We consider a canonical target tracking problem \cite{bar2004estimation}. Consider a system with linear-Gaussian dynamics, where the state of the system is given by the tracked target's position along the x- and y-axis and its corresponding velocities, namely $\mathbf{x}_k=[x,\dot{x},y,\dot{y}]^T$. At the $k$-th time instance, this state evolves according to
\begin{equation}
\mathbf{x}_{k+1}=\mathbf{A} \mathbf{x}_k + \mathbf{w}_k,
\end{equation}
where the process noise is Gaussian distributed according to $\mathbf{w}_k \sim \mathcal{N}(\mathbf{0},\mathbf{Q})$, with covariance matrix $\mathbf{Q}$ given by \cite{bar2004estimation}
\begin{align}
\mathbf{Q}=
\begin{bmatrix}
\frac{\Delta^3}{3} & \frac{\Delta^2}{2} & 0 & 0\\
\frac{\Delta^2}{2} & \Delta & 0 & 0\\
0 & 0 & \frac{\Delta^3}{3} & \frac{\Delta^2}{2}\\
0 & 0 & \frac{\Delta^2}{2} & \Delta
\end{bmatrix},
\end{align}
where $\Delta$ is the sampling interval of the system. The system dynamics follow a nearly-constant velocity model with
\begin{align}
\mathbf{A}=
\begin{bmatrix}
1 & \Delta & 0 & 0\\
0 & 1 & 0 & 0\\
0 & 0 & 1 & \Delta \\
0 & 0 & 0 & 1
\end{bmatrix}.
\end{align}
A network of agents is deployed to estimate the trajectory of the target. Heterogeneous measurements are considered, with ranging (Time-Of-Arrival) and bearing (Direction-Of-Arrival) nodes being considered. These two types of nodes acquire target measurements using the following model 
\begin{align}\label{eq:measurement_model_num}
z^{i}_{k}=
\begin{cases}
\sqrt{(x_k - x^i)^2 + (y_k - y^i)^2} +v_{k}^{i,r}  &\text{if TOA}\\
\arctan\bigl(\frac{x_k-x^i}{y_k-y^i} \bigr)+v_{k}^{i,\theta} &\text{if DOA}
\end{cases}
\end{align}
where $v_{k}^{i,r} \sim \mathcal{N}(0,\sigma_r)$ and $v_{k}^{i,\theta} \sim \mathcal{N}(0,\sigma_\theta)$. The non-linearity of the measurement model is addressed following the usual extended Kalman filter procedure. To estimate the target's state, the nodes use the proposed filter (Algorithm \ref{alg:algorithm}). A map of the environment is shown in Fig. \ref{fig:map}.

\begin{figure}[t]
	\centering
	\includegraphics[scale=1]{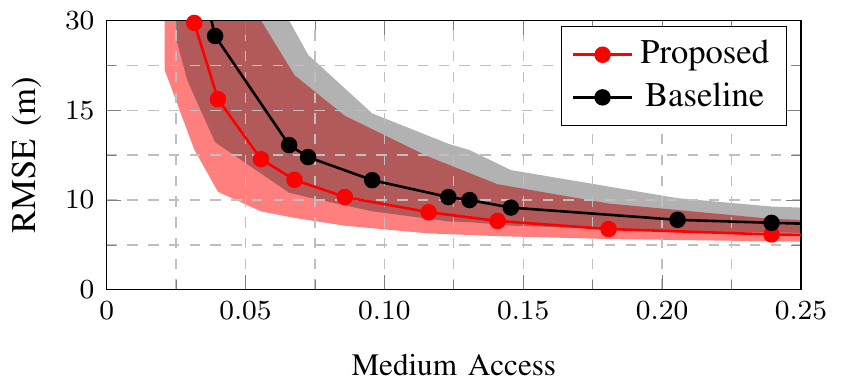} 			
\vspace*{-0.3in}
\caption{Operating region of the VoI filter. Solid lines correspond to the asymptotic values averaged over the network. Shaded interval corresponds to the range between the best and worst performing sensors in the network. The proposed method outperforms the VoI baseline.}
	\label{fig:region}
\vspace*{0.1in}
	\centering
	\includegraphics[scale=1]{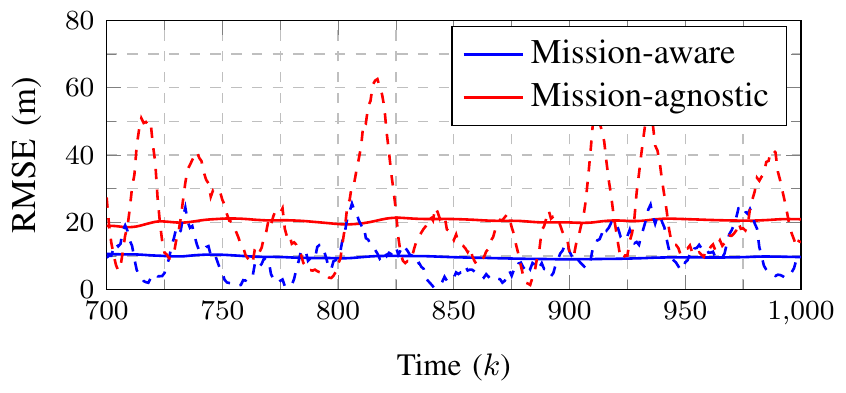} 			
\vspace*{-0.3in}
\caption{Comparison performance between a mission-aware filter and its agnostic counterpart. Solid lines correspond to running average errors, while the instantaneous error is shown by the dashed lines. The mission requirement is for the DOA node at the top left corner (see Fig. \ref{fig:map}) to have an average estimation error (solid lines) under $10$ m. Only the mission-aware filter satisfies this requirement, with an average error of $9.72$ m.}
	\label{fig:mission}
\end{figure}

We start by evaluating the region of operation of the proposed filter, as shown in Fig. \ref{fig:region}. This corresponds to the trade-off between error and medium access, obtained by sweeping over a range of possible censoring values $\gamma$. As the threshold $\gamma$ increases in value, the nodes produce more censoring decisions. Hence, reducing the number of transmission, but unavoidably leading to an increase in the overall estimation error. However, as previously discussed, a large reduction of transmission can be obtained with minimal effect in the estimation error. We compare the algorithm introduced in this work, with the baseline VoI algorithm introduced in our previous work \cite{calvo2022distributed}. The baseline filter utilizes a similar (mission-agnostic) VoI censoring mechanism and LogOP aggregation, but is formulated in a simpler first-order manner, without relying on an augmented Lagrangian and ADMM formulation. We observe that the ADMM approach introduced in this work allows us to obtain a more efficient region of operation for the filter. With a characterized region of operation, a choice in trade-off between estimation error and transmissions can be made, which in the following we set to the knee of the curve.

As discussed in Section \ref{sec:mission_aware}, the proposed algorithm allows for mission-aware operation. In this case, we evaluate the same scenario highlighted in Fig. \ref{fig:map}, but now we require the DOA agent situated at the top left corner to have an average estimation error under $10$ m. The resulting estimation error is shown in Fig. \ref{fig:mission}. For this mode of operation, the mission-agnostic form of the filter does not satisfy the requirement, which is not a fault of the algorithm, as it is simply not specified. In contrast, the mission-aware version of the filter succeeds in satisfying the average requirement.

\begin{figure}[t]
	\centering
	\includegraphics[scale=1]{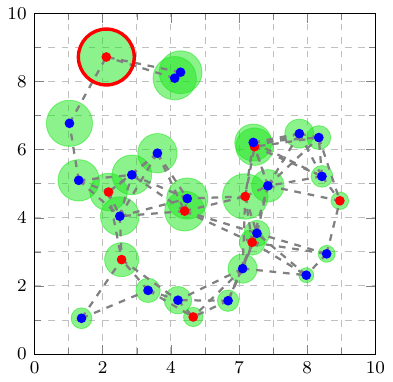} 			
	\includegraphics[scale=1]{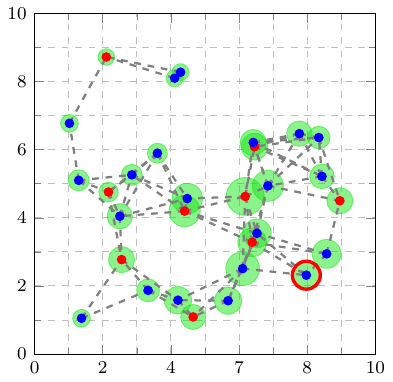} 			
\caption{Mission requirement for nodes highlighted by red circles is to have an estimation error $\leq 10$ m. Green circles correspond to amount of network transmissions (radius of $2$ corresponds to 100\% transmission usage).}
	\label{fig:flows}
\end{figure}

To better understand the network behavior induced by the mission-aware filter, we show in Fig. \ref{fig:flows} how information flows among nodes. In this figure, we compare the same mission but specified for two distinct agents. The left figure corresponds to the behavior discussed previously in Fig. \ref{fig:mission}. For this case, the average transmission rate across the network is of $27\%$, with the highlighted node transmitting $41$\% of the time. In contrast, specifying the same mission to the node shown on the right requires an average transmission of $18$\% across the network, with the highlighted node transmitting in $19\%$ of the time instances. In this scenario, the left specification is more restrictive, as it is placed on an isolated node that heavily relies on a small number of surrounding nodes to access the rest of the network. The node on the right is more densely connected, so when the mission changes to that node, the overall level of transmissions across the network are reduced, and as expected, the communication for the top left nodes is reduced dramatically. Importantly, regardless of the underlying network usage, both cases satisfy the mission requirement. Further, we can observe how certain nodes play an important role in the distributed estimation task, regardless of the mission requirement, e.g., the DOA node in the center of the network $(7.1,4.5)$ is both well-situated to sense the target and is mostly surrounded by TOA nodes, resulting in high value of information and consequently a relatively higher than average level of transmissions for both missions.

\section{Conclusions}

In this paper, we have introduced a new communication-efficient and mission-aware distributed filtering algorithm. By formulating the MAP estimation problem and having each agent sequentially solve a windowed version of the batch problem, we have more closely approximated the centralized solution. A distributed version of the problem has been achieved by the use of ADMM, wherein communication has been restricted to neighbors only. We have provided further communication reductions by the use of a mission-aware VoI censoring mechanism that provides communication-efficiency while ensuring that agents satisfy their local estimation requirements. Target tracking simulations have served to validate the viability of our approach.

\balance
\bibliographystyle{ieeetr}
\bibliography{bib}

\end{document}